\documentclass[11pt]{article}
\makeatletter
\newif\ifacl@finalcopy
\newif\ifacl@anonymize
\newif\ifacl@linenumbers
\newif\ifacl@pagenumbers
\newif\ifacl@hyperref
\acl@finalcopytrue
\acl@anonymizefalse
\acl@linenumbersfalse
\acl@pagenumberstrue
\acl@hyperreftrue
\usepackage{xcolor}

\ifacl@linenumbers
  \usepackage[switch,mathlines]{lineno}

  \newcount\cv@tmpc@ \newcount\cv@tmpc
  \def\fillzeros[#1]#2{\cv@tmpc@=#2\relax\ifnum\cv@tmpc@<0\cv@tmpc@=-\cv@tmpc@\fi
    \cv@tmpc=1 %
    \loop\ifnum\cv@tmpc@<10 \else \divide\cv@tmpc@ by 10 \advance\cv@tmpc by 1 \fi
      \ifnum\cv@tmpc@=10\relax\cv@tmpc@=11\relax\fi \ifnum\cv@tmpc@>10 \repeat
    \ifnum#2<0\advance\cv@tmpc1\relax-\fi
    \loop\ifnum\cv@tmpc<#1\relax0\advance\cv@tmpc1\relax\fi \ifnum\cv@tmpc<#1 \repeat
    \cv@tmpc@=#2\relax\ifnum\cv@tmpc@<0\cv@tmpc@=-\cv@tmpc@\fi \relax\the\cv@tmpc@}%
  
  \AtBeginDocument{\linenumbers}

  \usepackage{etoolbox} 

  \newcommand*\linenomathpatch[1]{%
    \expandafter\pretocmd\csname #1\endcsname {\linenomath}{}{}%
    \expandafter\pretocmd\csname #1*\endcsname {\linenomath}{}{}%
    \expandafter\apptocmd\csname end#1\endcsname {\endlinenomath}{}{}%
    \expandafter\apptocmd\csname end#1*\endcsname {\endlinenomath}{}{}%
  }
  \newcommand*\linenomathpatchAMS[1]{%
    \expandafter\pretocmd\csname #1\endcsname {\linenomathAMS}{}{}%
    \expandafter\pretocmd\csname #1*\endcsname {\linenomathAMS}{}{}%
    \expandafter\apptocmd\csname end#1\endcsname {\endlinenomath}{}{}%
    \expandafter\apptocmd\csname end#1*\endcsname {\endlinenomath}{}{}%
  }

  \expandafter\ifx\linenomath\linenomathWithnumbers
    \let\linenomathAMS\linenomathWithnumbers
    \patchcmd\linenomathAMS{\advance\postdisplaypenalty\linenopenalty}{}{}{}
  \else
    \let\linenomathAMS\linenomathNonumbers
  \fi

  \AtBeginDocument{%
    \linenomathpatch{equation}%
    \linenomathpatchAMS{gather}%
    \linenomathpatchAMS{multline}%
    \linenomathpatchAMS{align}%
    \linenomathpatchAMS{alignat}%
    \linenomathpatchAMS{flalign}%
  }
\else
  \newcommand{\@LN@col}[1]{}
  \newcommand{\@LN}[2]{}
  
\fi

\PassOptionsToPackage{a4paper,margin=2.5cm,heightrounded=true}{geometry}
\RequirePackage{geometry}

\newlength\titlebox
 \twocolumn 

\def\addcontentsline#1#2#3{}

\ifacl@pagenumbers
\else
\fi

\newcommand\outauthor{%
    \begin{tabular}[t]{c}
    \ifacl@anonymize
        \bfseries Anonymous ACL submission
    \else
        \bfseries\@author
    \fi
    \end{tabular}}

\AtBeginDocument{
\def\maketitle{\par
 \begingroup
   \def\thefootnote{\fnsymbol{footnote}}
   \twocolumn[\@maketitle]
   \@thanks
 \endgroup
 \setcounter{footnote}{0}
 \let\maketitle\relax
 \let\@maketitle\relax
 \gdef\@thanks{}\gdef\@author{}\gdef\@title{}\let\thanks\relax}
\def\@maketitle{\vbox to \titlebox{\hsize\textwidth
 \linewidth\hsize \vskip 0.125in minus 0.125in \centering
 {\Large\bfseries \@title \par} \vskip 0.2in plus 1fil minus 0.1in
 {\def\and{\unskip\enspace{\rmfamily and}\enspace}%
  \def\And{\end{tabular}\hss \egroup \hskip 1in plus 2fil
           \hbox to 0pt\bgroup\hss \begin{tabular}[t]{c}\bfseries}%
  \def\AND{\end{tabular}\hss\egroup \hfil\hfil\egroup
          \vskip 0.25in plus 1fil minus 0.125in
           \hbox to \linewidth\bgroup\large \hfil\hfil
             \hbox to 0pt\bgroup\hss \begin{tabular}[t]{c}\bfseries}
  \hbox to \linewidth\bgroup\large \hfil\hfil
    \hbox to 0pt\bgroup\hss
  \outauthor
   \hss\egroup
    \hfil\hfil\egroup}
  \vskip 0.3in plus 2fil minus 0.1in
}}
}

\renewenvironment{abstract}%
  {\begin{center}\large\textbf{\abstractname}\end{center}%
    \begin{list}{}%
      {\setlength{\rightmargin}{0.6cm}%
        \setlength{\leftmargin}{0.6cm}}%
      \item[]\ignorespaces%
      \@setsize\normalsize{12pt}\xpt\@xpt
  }%
  {\unskip\end{list}}

\RequirePackage{caption}
\DeclareCaptionFont{10pt}{\fontsize{10pt}{12pt}\selectfont}
\RequirePackage{natbib}
\renewcommand\cite{\citep}  

\def\thebibliography#1{\vskip\parskip%
\vskip\baselineskip%
\def\baselinestretch{1}%
\ifx\@currsize\normalsize\@normalsize\else\@currsize\fi%
\vskip-\parskip%
\vskip-\baselineskip%
\section*{References\@mkboth
 {References}{References}}\list
 {}{\setlength{\labelwidth}{0pt}\setlength{\leftmargin}{\parindent}
 \setlength{\itemindent}{-\parindent}}
 \def\newblock{\hskip .11em plus .33em minus -.07em}
 \sloppy\clubpenalty4000\widowpenalty4000
 \sfcode`\.=1000\relax}

\def\thesourcebibliography#1{\vskip\parskip%
\vskip\baselineskip%
\def\baselinestretch{1}%
\ifx\@currsize\normalsize\@normalsize\else\@currsize\fi%
\vskip-\parskip%
\vskip-\baselineskip%
\section*{Sources of Attested Examples\@mkboth
 {Sources of Attested Examples}{Sources of Attested Examples}}\list
 {}{\setlength{\labelwidth}{0pt}\setlength{\leftmargin}{\parindent}
 \setlength{\itemindent}{-\parindent}}
 \def\newblock{\hskip .11em plus .33em minus -.07em}
 \sloppy\clubpenalty4000\widowpenalty4000
 \sfcode`\.=1000\relax}

\def\section{\@startsection {section}{1}{\z@}{-2.0ex plus
    -0.5ex minus -.2ex}{1.5ex plus 0.3ex minus .2ex}{\large\bfseries\raggedright}}
\def\subsection{\@startsection{subsection}{2}{\z@}{-1.8ex plus
    -0.5ex minus -.2ex}{0.8ex plus .2ex}{\normalsize\bfseries\raggedright}}
\def\subsubsection{\@startsection{subsubsection}{3}{\z@}{-1.5ex plus
   -0.5ex minus -.2ex}{0.5ex plus .2ex}{\normalsize\bfseries\raggedright}}
\def\paragraph{\@startsection{paragraph}{4}{\z@}{1.5ex plus
   0.5ex minus .2ex}{-1em}{\normalsize\bfseries}}
\def\subparagraph{\@startsection{subparagraph}{5}{\parindent}{1.5ex plus
   0.5ex minus .2ex}{-1em}{\normalsize\bfseries}}

\skip\footins 9pt plus 4pt minus 2pt
\def\footnoterule{\kern-3pt \hrule width 5pc \kern 2.6pt }
\labelwidth\leftmargini\advance\labelwidth-\labelsep \labelsep 5pt

\def\@listi{\leftmargin\leftmargini}
\def\@listii{\leftmargin\leftmarginii
   \labelwidth\leftmarginii\advance\labelwidth-\labelsep
   \topsep 2pt plus 1pt minus 0.5pt
   \parsep 1pt plus 0.5pt minus 0.5pt
   \itemsep \parsep}
\def\@listiii{\leftmargin\leftmarginiii
    \labelwidth\leftmarginiii\advance\labelwidth-\labelsep
    \topsep 1pt plus 0.5pt minus 0.5pt
    \parsep \z@ \partopsep 0.5pt plus 0pt minus 0.5pt
    \itemsep \topsep}
\def\@listiv{\leftmargin\leftmarginiv
     \labelwidth\leftmarginiv\advance\labelwidth-\labelsep}
\def\@listv{\leftmargin\leftmarginv
     \labelwidth\leftmarginv\advance\labelwidth-\labelsep}
\def\@listvi{\leftmargin\leftmarginvi
     \labelwidth\leftmarginvi\advance\labelwidth-\labelsep}

\belowdisplayskip \abovedisplayskip
\def\@normalsize{\@setsize\normalsize{11pt}\xpt\@xpt}
\def\small{\@setsize\small{10pt}\ixpt\@ixpt}
\def\footnotesize{\@setsize\footnotesize{10pt}\ixpt\@ixpt}
\def\scriptsize{\@setsize\scriptsize{8pt}\viipt\@viipt}
\def\tiny{\@setsize\tiny{7pt}\vipt\@vipt}
\def\large{\@setsize\large{14pt}\xiipt\@xiipt}
\def\Large{\@setsize\Large{16pt}\xivpt\@xivpt}
\def\LARGE{\@setsize\LARGE{20pt}\xviipt\@xviipt}
\def\huge{\@setsize\huge{23pt}\xxpt\@xxpt}
\def\Huge{\@setsize\Huge{28pt}\xxvpt\@xxvpt}

\ifacl@hyperref
  \PassOptionsToPackage{breaklinks}{hyperref}
  \RequirePackage{hyperref}
  \definecolor{darkblue}{rgb}{0, 0, 0.5}
  \hypersetup{colorlinks=true, citecolor=darkblue, linkcolor=darkblue, urlcolor=darkblue}
\else
  \def\href#1#2{{#2}}
  \usepackage{url}
\fi

\makeatother
\usepackage{times}
\usepackage{latexsym}
\usepackage[T1]{fontenc}
\usepackage[utf8]{inputenc}
\usepackage{booktabs,array,graphicx}
\usepackage{amsmath,amssymb}
\usepackage{microtype}
\usepackage{tikz}
\usetikzlibrary{arrows.meta,positioning}
\newcommand{\ArtifactAddress}{https://anonymous.4open.science/r/LEAST-LM-release-D5C1/README.md}
\newcommand{\ArtifactURL}{\expandafter\url\expandafter{\ArtifactAddress}}
\title{Shared Stopping Decisions Change Answers\\in HQQ Cache Quantization}
\author{
  Seunghui Jwa\textsuperscript{1} \quad
  Minsu Oh\textsuperscript{1} \quad
  Chanjun Park\textsuperscript{2} \quad
  Yeo-Chan Yoon\textsuperscript{1} \\
  \textsuperscript{1}Department of Artificial Intelligence, Jeju National University \\
  \textsuperscript{2}School of Software, Soongsil University
}
\hypersetup{
  pdfauthor={Seunghui Jwa, Minsu Oh, Chanjun Park, Yeo-Chan Yoon},
  pdftitle={Shared Stopping Decisions Change Answers in HQQ Cache Quantization}
}
\date{}
\begin{document}
\maketitle
\newcommand{\DevTwoText}{404}
\newcommand{\DevFourText}{227}
\newcommand{\FreshText}{170}
\newcommand{\FreshPass}{16}
\newcommand{\FreshNumericFlip}{8}
\newcommand{\FreshValue}{9}
\newcommand{\FreshResolved}{246}
\newcommand{\LocalMs}{12.9050}
\newcommand{\NativeMs}{8.3345}
\newcommand{\FixedFourMs}{3.1205}
\newcommand{\FixedTwentyMs}{7.8219}
\begin{abstract}
Language-model systems batch questions for throughput, but unrelated questions
should not change a target's answer when its input and numerical execution
are fixed. We study compression of the key and value cache, which stores
attention representations reused during generation. With request-local groups,
Transformers' Half-Quadratic Quantization (HQQ) backend updates compression
parameters separately but uses a shared average error to decide when all updates stop.
Replacing only the question batched with the target changes four-bit HQQ
answers in \FreshText/384 test comparisons across two models. Replaying the
other execution's update counts reproduces its complete answer and cache
fingerprints in every changed pair, in both directions. Computing the stopping
mean in FP32 reduces cache differences but leaves answer changes.
Native HQQ also changes confirmed numerical correctness in eight arithmetic pairs.
Fixed iterations and request-local stopping remove observed companion
dependence under matched controls. Request-local stopping remains sensitive to synthetic
padding changes at the tensor level. Fixing the original iteration budget removes this decision path
without tuning. Neither repair has an established
quality advantage, and natural rebatching still changes answers.
Request-independence audits must cover stopping decisions as well as
quantization groups.
\end{abstract}

\section{Introduction}
Serving systems batch unrelated questions for efficiency, but a target's answer
should not depend on other questions in its batch, called companions. Beyond
known floating-point effects \citep{yuan2025numerical,vllm2026batch}, we test for dependence from cache
compression with target input, positions, attention mask, and padded batch shape fixed.

The key and value (KV) cache stores attention representations reused during
generation. Quantization stores them as low-bit codes with reconstruction
parameters. Half-Quadratic Quantization (HQQ) iteratively fits these parameters
\citep{badri2023hqq} and is a cache backend in the Transformers model execution
library \citep{wolf2020transformers,hf2026cacheguide}. Although quantization
groups and updates stay within each request, HQQ averages reconstruction error
across requests to decide when all updates stop. Companions can change the target's update count.

Development observations inform the repair baselines; a separate test set
evaluates their behavior and the causal role of stopping counts.
Replacing only the companion changes four-bit HQQ answers in \FreshText/384
comparisons of 192 test questions across two models. To test whether update
counts cause these changes, we replay each compression step's recorded count.
Under matched computation, all \FreshText{} changed pairs reproduce the
corresponding complete answer and cache fingerprints in both directions.

Text changes need not change correctness. We compare final numerical answers with references, separately from
output-format scores.

To remove companion dependence, we compare fixed iterations, initialization
without optimization, and request-local stopping. All restore consistency
under matched controls. We also measure their effects on answer quality and compression cost.\footnote{\href{\ArtifactAddress}{Code and verification records} (Appendix~\ref{app:artifact}).}

\section{Related Work}
Greedy decoding alone does not ensure repeatability
\citep{atil-etal-2025-non}. Batch-dependent numerical execution can change
answers \citep{yuan2025numerical,he2025determinism}; we match target computation
to isolate a separate decision shared across requests.

KIVI and KVQuant compress KV caches while preserving answer quality
\citep{liu2024kivi,hooper2024kvquant}. We test whether answers also remain
consistent when companions change.

Requests can interfere through shared expert capacity
\citep{hayes2024buffer} or dynamic quantization statistics, as Quantamination
shows \citep{foerster2026quantamination}.
Shared scales across token positions can also let future tokens affect
earlier attention outputs \citep{mishra2026leaky}. We test whether separating
quantization parameters and updates suffices for request independence.

HQQ iteratively fits compression parameters without calibration data
\citep{badri2023hqq}. In Transformers' HQQ cache integration, groups and
updates remain request-local, but the legacy optimizer averages errors across
requests to decide when all updates stop \citep{hqq2026source}.
Companions can thus change update counts. We reproduce changed answers by
replaying counts under matched computation and verify fixed or per-request
stopping as repairs. We check independence of termination as well as quantization parameters.

\section{Local Updates with a Shared Stop}
\label{sec:mechanism}
\begin{figure}[ht]
\centering
\begin{tikzpicture}[x=1cm,y=1cm,>=Latex,font=\small]
\node at (3.65,1.75) {Same target input and local update trajectory};
\draw[->,thick] (0,1.12) -- (7.3,1.12);
\node[above] at (.25,1.12) {$z_g(0)$};
\fill[blue!75!black] (2.1,1.12) circle (2pt);
\node[above,text=blue!75!black] at (2.1,1.12) {$J_A$};
\fill[orange!85!black] (5.2,1.12) circle (2pt);
\node[above,text=orange!85!black] at (5.2,1.12) {$J_B$};
\draw[->,blue!75!black] (2.1,1.02) -- (2.1,.55);
\draw[->,orange!85!black] (5.2,1.02) -- (5.2,.55);
\node[text=blue!75!black] at (2.1,.32) {Cache A};
\node[text=orange!85!black] at (5.2,.32) {Cache B};
\draw[green!40!black,thick] (3.55,.98) -- (3.55,1.26);
\node[above,text=green!40!black] at (3.55,1.12) {$k$};
\node[below,align=center,text=green!40!black] at (3.65,-.05)
 {Fixed $k$: both companions select the same endpoint};
\end{tikzpicture}
\caption{Schematic of initial compression. Different companions can select
counts $J_A,J_B$ on the same target trajectory, yielding different caches.
A fixed count $k$ removes this endpoint dependence.}
\label{fig:mechanism}
\end{figure}
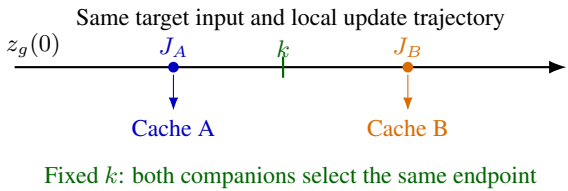

Our cache interventions change compression during inference while keeping
model weights and greedy decoding fixed. Let $X_r$ be request $r$'s cache
tensor and $z_g(j)$ the parameters of a group $g$ within that request after
$j$ updates. With initialization $I_g$ and update $F_{g,j}$,
\begin{equation}
 z_g(0)=I_g(X_r),\quad z_g(j+1)=F_{g,j}(X_r,z_g(j)).
\end{equation}
The predetermined penalty schedule is indexed by $j$ in $F_{g,j}$.
If a call with $b$ requests stops at $J(X_1,\ldots,X_b)$, reconstruction $R_g$
returns $\widehat X_g=R_g(X_r,z_g(J))$.
Fixing $X_r$ fixes the trajectory, but another request can change $J$.
Figure~\ref{fig:mechanism} shows this endpoint selection. Different counts affect the cache only if their reconstructions differ.

The pinned legacy optimizer reconstructs with the current parameters, then
updates zero points using group-local means. It tests the earlier
reconstruction's tensor-wide mean absolute error for strict improvement,
retaining the updated parameters even when the test fails
\citep{hqq2026source}. We preserve this ordering, the 20-update cap, and FP16
optimization of BF16 inputs. Fixed $k$ updates remove the shared test;
disabling optimization keeps initialization. Request-local stopping applies
the native test per request and freezes finished requests with masks.
Appendix~\ref{app:preservation} gives the update and test order.

These repairs require local grouping, updates, reconstruction, and consistent
numerical execution at every cache update. HQQ repeatedly recompresses
reconstructed history with newly accumulated tokens. The first answer token
precedes any compressed-cache read, so we check subsequent decoding and rolling
recompression. Appendix~\ref{app:preservation} states the preservation condition.

\section{Experiments}
\label{sec:experiments}
\begin{figure*}[t]
\centering
\begin{tikzpicture}[x=1cm,y=1cm,>=Latex,font=\small,
 every node/.style={inner sep=2pt},
 flow/.style={->,line width=.65pt,draw=black!60},
 laneA/.style={draw=blue!55!black,fill=blue!5,rounded corners=2pt,
   minimum height=.56cm,align=center},
 laneB/.style={draw=orange!70!black,fill=orange!8,rounded corners=2pt,
   minimum height=.56cm,align=center},
 title/.style={font=\small\bfseries,align=center},
 note/.style={font=\small,align=center,text=black!80}]
\node[title] at (1.6,2.10) {1\quad Match the target};
\node[title] at (5.55,2.10) {2\quad Record native runs};
\node[title] at (9.65,2.10) {3\quad Apply other counts};
\node[title] at (13.75,2.10) {4\quad Test full transfer};

\node[laneA,text width=2.80cm] (inputA) at (1.6,1.50)
 {Run A: $(\mathbf q,a)$};
\node[laneB,text width=2.80cm] (inputB) at (1.6,.70)
 {Run B: $(\mathbf q,b)$};
\node[laneA,text width=3.05cm] (nativeA) at (5.55,1.50)
 {$\mathbf J_A$, cache A, answer A};
\node[laneB,text width=3.05cm] (nativeB) at (5.55,.70)
 {$\mathbf J_B$, cache B, answer B};
\node[laneA,text width=2.90cm] (replayA) at (9.65,1.50)
 {Run A using $\mathbf J_B$};
\node[laneB,text width=2.90cm] (replayB) at (9.65,.70)
 {Run B using $\mathbf J_A$};
\node[laneA,text width=3.10cm] (checkA) at (13.75,1.50)
 {Compare with native B};
\node[laneB,text width=3.10cm] (checkB) at (13.75,.70)
 {Compare with native A};

\draw[flow] (inputA) -- (nativeA);
\draw[flow] (inputB) -- (nativeB);
\draw[flow,draw=blue!65!black] (nativeA.east) -- (replayB.west);
\draw[flow,draw=orange!80!black,preaction={draw=white,line width=2.7pt}]
 (nativeB.east) -- (replayA.west);
\draw[flow] (replayA) -- (checkA);
\draw[flow] (replayB) -- (checkB);

\node[note,text width=3.10cm] at (1.6,.03)
 {Same initial state\\and numerical layout};
\node[note,text width=3.20cm] at (5.55,.03)
 {All native\\text-changing pairs};
\node[note,text width=3.20cm] at (9.65,.03)
 {Same prompt\\and update arithmetic};
\node[note,text width=3.30cm] at (13.75,.03)
 {Cache hashes\\and full output tokens};
\end{tikzpicture}
\caption{Causal verification on the test set for target $q$ with companions $a,b$.
$\mathbf J_A,\mathbf J_B$ record update counts for each layer and key/value
compression call.
Tracing, own-count replay and restoration must reproduce native execution.
Cross-run replay changes only counts, recomputing caches and answers.
Full transfer requires all raw/reconstructed cache hashes and output tokens
to match the opposite run through EOS or the cap, with no missing counts.}
\label{fig:verification-pipeline}
\end{figure*}
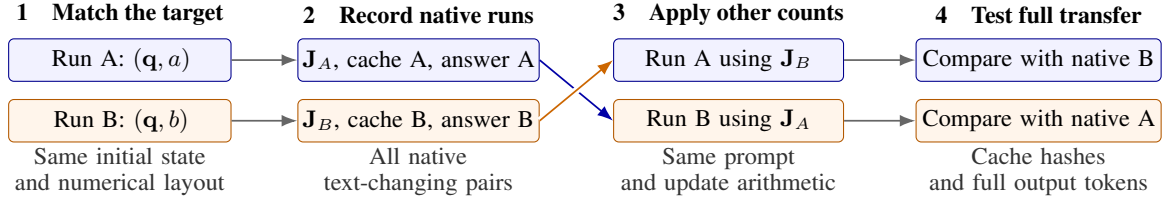

Figure~\ref{fig:verification-pipeline} summarizes the controlled comparison
and causal checks.
\paragraph{Controlled comparison.}
We test whether replacing a companion changes the target's answer through
shared stopping. Each pair fixes the model and target $q$ in batches $(q,a)$
and $(q,b)$. Target token IDs, positions, attention mask, and padded batch
shape are identical, with greedy decoding. Both runs use the same explicit
mask implementation. Target first logits and raw cache inputs at initial
compression must agree. With request separation, target continuations should match through their
end-of-sequence (EOS) token or output cap.

We use development questions to investigate dependence and choose baselines;
a separate test set evaluates repairs and the causal role of stopping counts. The development set contains 32 questions
from each of six public QA, arithmetic, and instruction datasets, totaling 192
questions. Five models give 960 model/question pairs. The test set contains 128
arithmetic questions from GSM8K's official test split \citep{cobbe2021gsm8k}
and 64 document-QA questions from HotpotQA's official validation split with
distractor documents \citep{yang2018hotpotqa}.
Llama-3.1-8B-Instruct \citep{grattafiori2024llama3} and Qwen3-8B with thinking
disabled \citep{yang2025qwen3} each answer all 192 test targets with two fixed
companions, giving 384 pairs.
The test set excludes earlier source IDs and normalized questions, including
those in the development set. HotpotQA also excludes document-title overlap
within the test set and with development inputs.
Appendix~\ref{app:settings} gives models, prompts, companion selection,
revisions, and sampling restrictions.

For the test-set comparison, model weights remain BF16. HQQ is opt-in;
we select recommended key/value axes 1/1, versus constructor defaults 0/0
\citep{hf2026cacheguide}. Shared stopping couples requests within the same
compression call. Four-bit groups span 64 channels within tokens;
residual capacity is 128 tokens. QA and arithmetic output caps
are 64 and 512 tokens. Seven methods yield 5,376 baseline answers: BF16,
native HQQ, no optimization, fixed 4, fixed 20, request-local stopping, and
Optimum Quanto, an external affine cache-quantization baseline
\citep{huggingface2025quanto}. Favorable Qwen development observations motivated
including fixed 4 before test generation, without a prespecified optimality
criterion (Appendix~\ref{app:budget-choice}).

\paragraph{Intervention and checks.}
We test whether the changed answer follows the compression update counts.
For all \FreshText{} changed test pairs, we keep the questions fixed and
replay the other run's recorded counts at each layer and key/value compression
call. Replay should then reproduce that run's answer. Both directions are tested.
Tracing, replay of each run's own counts, and restored native execution must
preserve its original outputs. Full transfer requires the
opposite recorded raw and reconstructed target cache fingerprints and complete
token/EOS sequence in both directions, with a recorded count available for
every compression call before the target ends. Later target inputs can diverge once cache differences affect
generation; replay must transfer those events too. Replaying recorded counts is diagnostic, not an online compression policy.

\paragraph{Outcome definitions.}
Text changes compare decoded strings after removing special tokens and outer
whitespace; token changes compare complete generated token-ID sequences,
including EOS. Official pass uses the pinned GSM8K marker parser, HotpotQA
answer exact match, or the development task's scorer (Appendix~\ref{app:scoring}).
A separate extractor, frozen before generation, reads explicit final numbers
without consulting the reference. Missing or conflicting assertions remain
unresolved. We distinguish changed numerical values from changed correctness
against the reference; neither verifies the reasoning steps.

\section{Results and Repair Boundaries}
\label{sec:results}
\paragraph{Development and test results.}
On the development set, the five-model comparison tests dependence across models
and tasks: native HQQ
changes \DevTwoText/960 texts at two bits and \DevFourText/960 at four bits.
Four-bit official-score changes occur only in math and instruction following
(Tables~\ref{tab:models}--\ref{tab:datasets}). On the separate test set of 128 GSM8K math and 64 HotpotQA
document-QA questions, two models evaluate the causal role of stopping counts
and compare repairs. Native HQQ changes \FreshText/384 texts;
BF16, no optimization, fixed 4/20, request-local stopping and Quanto each
give 0/384 text or token changes under matched controls.
\begin{table}[t]
\centering
\small
\begin{tabular}{llrr}
\toprule
Model & Task & Text $\Delta$ & \shortstack{Numerical correctness\\$\Delta$/resolved} \\
\midrule
Llama & Math & 80/128 & 6/122 \\
Qwen & Math & 89/128 & 2/124 \\
Both & Math & 169/256 & 8/246 \\
\midrule
Llama & QA & 0/64 & -- \\
Qwen & QA & 1/64 & -- \\
\bottomrule
\end{tabular}
\caption{Test-set answer changes under four-bit HQQ companion replacement. Text $\Delta$ counts changed strings. Numerical correctness $\Delta$/resolved counts correct-to-incorrect or reverse changes over pairs where both final numbers can be extracted, without requiring an answer marker. Both sums the two math rows; the models reuse questions.}
\label{tab:prospective}
\end{table}

Across the two models, 169/256 math pairs change text
(Table~\ref{tab:prospective}). Both final numbers can be extracted in
\FreshResolved{} pairs; five change from correct to incorrect and three
from incorrect to correct. Appendix~\ref{app:quality} lists all eight pairs
and their asserted numbers. Numerical correctness
compares the extracted final number with the reference without requiring an
answer marker. In ten pairs, at least one final number remains unresolved; these are not
counted as known numerical errors.

\paragraph{Reproducible changes under count intervention.}
In Figure~\ref{fig:verification-pipeline}, questions stay fixed while each
compression uses the other run's recorded update count. All \FreshText{}
changed pairs then reproduce that run's complete tokens and cache hashes,
in both directions. Under matched computation, the answer change is
reproducibly linked to the per-compression count sequence.
Across all 384 pairs, an FP32 stopping mean reduces initial cache changes
from 10,543 to 614 of 26,112 calls. Across all 256 math pairs, answers still
change in 129 (native 169), including 23 with initially identical caches
that diverge during recompression. Updates remain FP16 and stopping remains
shared (Appendix~\ref{app:stop-statistics}).

\paragraph{Why fewer QA answers change.}
QA requests short document-based answers; math requests reasoning.
Exploratory median lengths are 5 tokens for QA in both models, versus 202
and 266.5 for math. Of 169 math changes, 94 begin after token 64; 75/256
math pairs still differ within 64 tokens, versus 1/128 QA pairs.
All 128 QA pairs have different initial reconstructed caches.
Longer outputs offer more opportunities for divergence, but prompts, tasks
and EOS also differ; length is not isolated as the cause
(Appendix~\ref{app:task-difference}).

\paragraph{Repairs and their scope.}
To retain iterative fitting without tuning, use the original fixed
20-update budget. On 96 actual initial-KV tensors, geometric-mean compression
time is 6.22 ms versus 6.35 for native HQQ; fixed 20 is slower for Llama QA.
No optimization costs 0.52 ms and also restores matched consistency.
No quality advantage is established.
In a synthetic padding control, request-local stopping changes attended
reconstructions in 58/96 comparisons,
versus 0/96 for fixed 20 (Appendices~\ref{app:stop-statistics}--\ref{app:cost}).

\paragraph{Natural batch configurations.}
We also rebatch all 192 test questions in groups of four using natural
padding and scaled dot-product attention (SDPA). Six shuffles are compared
with each method's source-order answers (Table~\ref{tab:natural-summary}).
Fixed 20 reduces text changes by 8.4 percentage points on average,
but numerical-correctness changes remain similar to native HQQ and more
frequent than BF16. This comparison changes both stopping and update counts,
and mixes their effects with padding and numerical execution; it does not
isolate a shared-stopping effect or establish improved answer correctness.
All nine development methods and the seed-level test results are in
Appendix~\ref{app:rebatch-tasks}.
\begin{table}[t]
\centering\small
\begin{tabular}{lrr}
\toprule
Method & Text changes & Correctness changes \\
\midrule
BF16 & 36.1\% & 17/1,484 \\
Native HQQ & 54.9\% & 64/1,501 \\
Fixed 20 & 46.6\% & 62/1,496 \\
\bottomrule
\end{tabular}
\caption{Natural test-set rebatching across six shuffle seeds. Text: mean over 384 model/question pairs per seed. Correctness: final-number correctness changes / pairs with both numbers resolved, summed across seeds for arithmetic only. These reuse the same questions; the denominators are not independent samples.}
\label{tab:natural-summary}
\end{table}

\paragraph{Extension to weight compression.}
Joint versus separate Q/K/V weight compression also changes answers across
three models. All 300 matrix count interventions reproduce the opposite
compressed values and metadata; 500 fixed-budget checks preserve them across
partitions. This extends the mechanism within HQQ (Appendix~\ref{app:weights}).

\section{Conclusion}
Local quantization groups do not ensure request separation when termination
is shared. Count replay under matched computation connects shared stopping
to complete answers with HQQ caches. Fixed iterations repair this dependence.
Request-local stopping preserves independent compression's stopping behavior
on the same padded input, with no established practical advantage over fixed
iterations.
\clearpage
\section*{Limitations}
The controlled cache study uses two-request batches and a pinned HQQ legacy
optimizer in Transformers, with requests entering the same compression call. Requests sharing a GPU need not share a compression call, so GPU sharing
alone does not imply this dependence. The study does not establish how often
this configuration is deployed, failure in every HQQ backend, or information
disclosure.
The weight study extends the evaluation beyond KV caches, but uses the same
HQQ lineage rather than a second independent quantization mechanism. Its
controlled BF16 reconstruction does not evaluate an entire TorchAO runtime
or actual int4 inference kernels.

The causal intervention and the stored-record audit have different scopes.
Exact causal transfer is established for the 170 changed native-HQQ test
pairs and covers recorded runs through EOS or the fixed generation cap. The offline audit
checks the stored cache fingerprints, computed from contiguous FP32-converted
tensor representations; it does not independently recompute hidden states.

The preservation argument, in turn, assumes matched numerical execution.
Natural rebatching can change shapes and kernels, so the argument does not
extend to arbitrary batch configurations. The six-seed rebatching contrast
is conditional on the fixed test questions and two models, assumes symmetric
seed-level differences, and compares only BF16, native HQQ and fixed 20.
It does not establish a population-level accuracy or serving benefit.
The FP32 generation ablation covers all test arithmetic questions but retains
only 16 input-selected QA questions per model; it reuses test inputs rather
than providing a new independent accuracy evaluation.

Questions are reused across models and methods. Development questions are not
test-set accuracy measurements. Test-set exclusions of earlier source
identities and HotpotQA document titles constrain the sampling population;
they neither prove topic independence nor rule out pretraining exposure.
The three-model weight study excludes earlier question identities and hashes,
not document overlap. Its questions are shared across three models.
Descriptive bootstrap intervals are model/task specific and unadjusted for
multiple comparisons. Zero observed changes are not a zero-risk guarantee.
The final-number extractor covers explicit recognized assertions, not all
reasoning, units, or implicit answers. Unresolved outputs are retained.
In the three-model weight study, 200/256 Mistral arithmetic pairs have an
unresolved side, preventing a general accuracy conclusion for that model.

Actual-KV timings cover 96 initial-cache tensors from eight model/question
pairs and three layers, excluding rolling recompression and generation. Stored
cache bytes exclude model weights, workspace, and allocator reservation. In
addition, the CPU and bus remain shared during timing, limiting isolation
from other workloads. These scope and resource constraints preclude claims
of production throughput, service-level benefit, or hardware speed differences.
The measurements also do not establish general accuracy/cost superiority.

\section*{Ethical considerations}
We use public benchmark material and model checkpoints under their original
terms. No private user prompts, production traffic, or new human annotations
were collected.

AI assistance supported implementation, experiment orchestration, literature
checking, analysis, and writing. The authors personally reviewed all AI-assisted
work. The authors are responsible for the paper's claims and submission declarations.

\clearpage
\appendix
\section{Data, Models, and Execution}
\label{app:settings}
The development set contains 32 targets per dataset: document QA from
HotpotQA \citep{yang2018hotpotqa}, MuSiQue \citep{trivedi2022musique},
TriviaQA \citep{joshi2017triviaqa}, and SQuAD \citep{rajpurkar2016squad};
arithmetic from GSM8K \citep{cobbe2021gsm8k}; and verifiable instructions
from Instruction-Following Evaluation (IFEval; \citealp{zhou2023ifeval}).
The QA and arithmetic development examples use training partitions; IFEval
uses a development allocation from its public prompts. Each target has two
companions, giving 576 source requests. The selection seed is 609152.
These development results are not test accuracy estimates.

The five checkpoints are Llama-3.1-8B-Instruct
\citep{grattafiori2024llama3}, Qwen3-8B \citep{yang2025qwen3},
Mistral-7B-Instruct-v0.3 \citep{mistral2024v03}, Phi-4
\citep{abdin2024phi4}, and OLMo-2-0425-1B-Instruct
\citep{olmo2025oneb}. OLMo is a smaller comparison, not a matched-size model.
The respective revision prefixes are \texttt{0e9e39f249a1},
\texttt{b968826d9c46}, \texttt{c170c708c41d},
\texttt{2db69c1c3e91}, and \texttt{48d788eca847}; the artifact records full
revisions. All methods use the same checkpoint and tokenizer for a model.
Qwen thinking is disabled. We use greedy Transformers generation
\citep{wolf2020transformers}, BF16 model weights, and each model's EOS IDs.

Document QA prompts supply the selected documents and request a short answer
without explanation. Arithmetic prompts request reasoning followed by a
numerical answer after \texttt{\#\#\#\#}. IFEval preserves its supplied
instruction constraints. The input cap is 8,192 tokens, or 4,096 for OLMo;
overflow retains the prescribed equal head/tail portions. Only TriviaQA
targets are truncated: 15, 16, 18, 15, and 21 of 32 for the models listed
above. The corresponding companion counts are 35, 36, 38, 35, and 44 of 64.
Output caps are 64 for QA, 512 for arithmetic, and 1,024 for IFEval.
Our test set uses GSM8K's official test split and HotpotQA's official validation
split with distractor documents, with the exclusions stated in the main text.
Development and test denote their roles in our study, not a random split of a
single dataset. Development observations motivated including fixed four
updates; the test set evaluates the frozen baseline and the causal role of
stopping counts. Appendix~\ref{app:budget-choice} reports that choice's basis.
Models reuse questions within each study.

The standard path is an explicit HQQ-backed quantized cache with axes 1/1;
HQQ is an opt-in backend, not the default Quanto backend. The source archive
provides a minimal public-API example; the full collector additionally
implements the explicit-mask and causal controls described here.
The axes matter: for a contiguous cache of shape $[2,8,128,128]$ and group
size 64, HQQ's axis-0 reshape puts values from both requests in every one of
4,096 groups. Axis 1 puts each group within a request. This index-mapping
check explains why our isolation of stopping applies to axes 1/1; it is not
a generation result for the default axes 0/0.

The controlled cache study uses PyTorch 2.7.0+cu126 \citep{paszke2019pytorch},
Transformers 5.17.0,
HQQ 0.2.8.post1 at revision \texttt{d88a488ec8aa}, and Optimum Quanto 0.2.7.
HQQ uses key/value axes 1/1, group size 64, residual capacity 128, proximal exponent 0.7,
initial penalty 10, multiplier 1.01, and at most 20 updates. The native CUDA
optimizer casts cache inputs to FP16 for fitting; interventions preserve its
actual reductions, casts, and endpoint convention. Quanto uses four bits,
key/value axes 0/0, group size 64, and residual capacity 128 through the
Transformers Quanto cache adapter. These axis numbers follow each backend's
own conventions and do not specify a common grouping across HQQ and Quanto.

The development controlled study ran on H100; test-set generation
and tensor costs ran on RTX A6000, retaining batch sizes,
precision, prompts, and length caps. The controlled study uses an explicit
attention-mask path for every method. Included cases require complete records,
matching first target logits, and matching initial raw target cache inputs.
All 384 prespecified test-set model--question pairs are represented in the
final analysis, with the same targets and companions as the frozen allocation.
No pair was excluded or replaced, and each completed pair is counted once.
The hardware groups are not compared for speed.

\paragraph{Companions, padding, and termination.}
Test-set candidates are ranked by SHA256 of the selection seed 917503,
dataset, source ID, and normalized-question hash. After the reported overlap
exclusions, the first $3N$ eligible examples in each dataset supply $N$
targets followed by two distinct companions per target. Companions are from
the same dataset, are not reused as targets, and are fixed before any output
is observed. Each model uses its native chat template with generation prompt,
the fixed date string 15 Sep 2026, and the recorded thinking setting.
After prescribed truncation, both two-request batches are left-padded to the
maximum token length across $q,a,b$. Position IDs come from the cumulative
attention mask, with masked positions set to 1. A finished companion remains
in the fixed batch and receives padding tokens while the target continues;
its row is not removed from compression. Each recorded answer is clipped
at its own first EOS or output cap.

\paragraph{Prompts and scoring.}
\label{app:scoring}
The arithmetic template is ``Solve the following math problem. Explain your
reasoning, then put the final numerical answer after `\#\#\#\#'.'' followed by
the question. QA asks for the shortest answer without explanation, followed
by numbered documents, the question, and an answer prefix. Exact templates,
rendered prompts, input token IDs, and hashes are supplied with the records.

Official GSM8K scoring uses the upstream \texttt{extract\_answer} and
\texttt{is\_correct} functions at revision \texttt{3101c7d50724}. It matches
the first \texttt{\#\#\#\#} followed by one space and a signed sequence of
digits, decimal points, or commas, removes commas, and compares the resulting
string with the reference. Missing matches fail this strict score. Thus
\texttt{Answer: 6} or \texttt{\#\#\#\# \$6} can fail despite an asserted 6.
HotpotQA uses its answer EM/F1 functions at revision \texttt{3635853403a8},
with lowercasing and removal of articles, punctuation, and redundant whitespace.
Other development tasks use pinned upstream scoring functions;
IFEval pass means all prompt constraints pass its strict checks.

The separate numerical extractor accepts marked answers, boxed expressions,
explicit answer clauses, and responses consisting only of a number and an
accepted unit. It normalizes supported formatting and fractions and requires
all recognized assertions to agree. No recognized assertion or conflicting
values leave the answer unresolved. Only then is the value compared with the
reference, so extraction cannot choose the reference-matching candidate.
It does not check units or all reasoning. Output text comparison removes
special tokens and outer whitespace only; token-ID comparison retains EOS.

\paragraph{Choice of the fixed budget.}
\label{app:budget-choice}
Fixed 4 was a development-informed baseline, not the output of a prespecified
optimization or noninferiority rule. An earlier comparison reused all 192
development questions in source-order, shuffled and length-sorted batches of
four, testing no optimization and fixed 2/4/8/20 alongside native HQQ,
request-local stopping, BF16 and Quanto. Table~\ref{tab:budget-development}
reports each model and task without pooling their different metrics.
Qwen fixed 4 exceeded native by 1.04 percentage points on GSM8K's strict
score and 2.08 on IFEval, at lower compression cost. The corresponding
Llama GSM8K score fell by 1.04 points. Qwen MuSiQue F1 rose by 0.29 points
while exact match fell by 1.04 points. These mixed observations motivated
inclusion before test generation; no accuracy-loss tolerance or combined
quality/cost objective was specified. No claim that four is optimal follows.
No optimization and fixed 20 remain test comparators.
\begin{table*}[t]
\centering
\small
\begin{tabular}{llrrrrrr}
\toprule
Model & Method & GSM & IFEval & Hotpot & MuSiQue & SQuAD & TriviaQA \\
\midrule
Llama & BF16 & 73.96 & 83.33 & 87.92 & 34.32 & 85.72 & 91.53 \\
Llama & Native HQQ & 66.67 & 81.25 & 87.92 & 34.32 & 85.72 & 91.53 \\
Llama & No optimization & 67.71 & 82.29 & 87.92 & 34.32 & 85.72 & 91.53 \\
Llama & Fixed 2 & 68.75 & 83.33 & 87.92 & 33.98 & 85.72 & 91.53 \\
Llama & Fixed 4 & 65.62 & 81.25 & 87.92 & 33.98 & 85.72 & 91.53 \\
Llama & Fixed 8 & 67.71 & 83.33 & 87.92 & 33.98 & 85.72 & 91.53 \\
Llama & Fixed 20 & 66.67 & 79.17 & 87.92 & 33.98 & 85.72 & 91.53 \\
Llama & Request-local & 67.71 & 78.12 & 87.92 & 33.98 & 85.72 & 91.53 \\
Llama & Quanto & 68.75 & 80.21 & 87.92 & 34.32 & 85.72 & 91.53 \\
Qwen & BF16 & 52.08 & 89.58 & 86.88 & 47.46 & 86.50 & 90.83 \\
Qwen & Native HQQ & 54.17 & 90.62 & 86.88 & 45.31 & 85.45 & 90.83 \\
Qwen & No optimization & 55.21 & 89.58 & 86.88 & 45.54 & 85.80 & 89.58 \\
Qwen & Fixed 2 & 53.12 & 89.58 & 86.88 & 44.50 & 85.80 & 90.83 \\
Qwen & Fixed 4 & 55.21 & 92.71 & 86.88 & 45.60 & 85.80 & 90.83 \\
Qwen & Fixed 8 & 52.08 & 87.50 & 86.88 & 43.75 & 85.80 & 90.83 \\
Qwen & Fixed 20 & 53.12 & 91.67 & 86.88 & 45.78 & 85.45 & 90.83 \\
Qwen & Request-local & 54.17 & 87.50 & 86.88 & 44.74 & 85.45 & 90.83 \\
Qwen & Quanto & 52.08 & 89.58 & 86.88 & 44.71 & 85.11 & 90.83 \\
\bottomrule
\end{tabular}
\caption{Development observations available before choosing test baselines. Each model/task averages 32 questions over three ordinary batch layouts, giving 96 outputs, not 96 independent questions. GSM and IFEval use strict correctness/compliance; QA columns use answer F1. Values are percentages. Tasks are not pooled into a selection objective. All budgets and unfavorable results are retained.}
\label{tab:budget-development}
\end{table*}

\begin{table}[htbp]
\centering
\small
\begin{tabular}{lrrrr}
\toprule
Model & 2b text & 2b pass & 4b text & 4b pass \\
\midrule
Llama 3.1 8B & 71 & 12 & 49 & 9 \\
Qwen3 8B & 126 & 8 & 40 & 3 \\
Mistral 7B & 80 & 2 & 53 & 4 \\
Phi-4 & 83 & 7 & 46 & 0 \\
OLMo 2 1B & 44 & 5 & 39 & 0 \\
\bottomrule
\end{tabular}
\caption{Native HQQ variation by model on the development set, with 192 question pairs per model. Pass denotes official task correctness or instruction compliance.}
\label{tab:models}
\end{table}
\begin{table}[htbp]
\centering
\small
\begin{tabular}{lrrrr}
\toprule
Dataset & 2b text & 2b pass & 4b text & 4b pass \\
\midrule
GSM8K & 142 & 6 & 92 & 12 \\
HotpotQA & 32 & 2 & 2 & 0 \\
IFEval & 146 & 22 & 121 & 4 \\
MuSiQue & 44 & 0 & 7 & 0 \\
SQuAD & 20 & 2 & 2 & 0 \\
TriviaQA & 20 & 2 & 3 & 0 \\
\bottomrule
\end{tabular}
\caption{Controlled native-HQQ changes by dataset on the development set, pooling five models over 32 shared targets (160 model--question pairs per row). Four-bit pass changes occur only in GSM8K and IFEval; text changes alone do not establish answer errors.}
\label{tab:datasets}
\end{table}

\section{Intervention and Preservation}
\label{app:preservation}
For both native runs, we record the number of updates at each layer and
compression call. To check that recording does not alter generation, we
require each run's output to match its output without recording. Replaying
the run's own counts must preserve its cache fingerprints and answer; otherwise,
the intervention would not encode the original stopping behavior. The cross
replay uses the corresponding other run's count without changing any update
formula. Restoration checks that instrumenting a model has not left persistent
state. Failure of a control or exhaustion of the donor schedule before the
target ends precludes a full-transfer claim.

The test-set records include 170 text-changing pairs, with all 170 passing
bidirectional transfer.

Full transfer requires matching the other run's recorded target cache
fingerprints and complete token sequence through EOS or the output cap.
An equal cache hash or equal first token alone is
insufficient. To check cache transfer, we store fingerprints of raw and
reconstructed target cache tensors at
every recorded event, including rolling recompression and masked padding
positions consumed by the optimizer. They hash contiguous FP32-converted
representations rather than storing every tensor. Thus offline verification
checks these stored fingerprints, token sequences, and replayed counts
without recomputing hidden states.

For update index $j=0,\ldots$, the legacy penalty is
$\beta_j=10(1.01)^j$. This predetermined schedule is part of $F_{g,j}$ in
Equation~(1), not a statistic of companion requests. The group's scale and
zero point initialize $z_g(0)$; scale stays fixed during the legacy proximal
updates. Each step reconstructs from the current parameters, computes its
local proximal residual, and updates zero points by a group-local mean.
The separate tensor-wide error reduction controls continuation. The replay
adapter preserves the step function, casts, penalty schedule, and retained
final update, replacing only the continuation decision with the recorded count.

Let $\pi_r(C)$ project cache state $C$ to all values needed for request $r$'s
future reads and writes, including quantization metadata, residual buffers,
positions, and index maps. A sufficient condition for cache transformation
$T$ to preserve separation is
\begin{equation}
 \pi_r(C)=\pi_r(C')\ \Longrightarrow\
 \pi_r(T(C))=\pi_r(T(C')).
\end{equation}
Assume equal target prefill state, matched numerical execution, a residual
trigger determined by target progress, and greedy decoding or common random
draws. Equal target cache state produces equal target logits and appended
values; applying the condition preserves equality after the next compression.
Induction gives equal continuations to EOS or the cap. Fixed counts satisfy
the endpoint part when initialization, updates, and reconstruction are local.
Request-local stopping applies the native test separately to each request's
padded cache tensor. Equivalence to isolated compression requires identical
tensor values, padding, and reduction arithmetic, including active-index
handling; it does not imply equality with an unpadded singleton run. This
argument does not guarantee separation under arbitrary kernel or layout changes.

\paragraph{Stopping rules in pseudocode.}
For each update, reconstruct using the current parameters, then update the
zero points. Compute the error from the earlier reconstruction. Native HQQ
compares the mean over all requests with the previous minimum; request-local
stopping compares one mean per request, including its padding positions.
Both retain the new zero points even when the error fails to improve.
Fixed $k$ uses the same update but omits the error test:
\begin{quote}\small\ttfamily
initialize; best = infinity\\
for j = 0, ..., limit - 1:\\
\hspace*{1em}reconstruct current values\\
\hspace*{1em}update unfrozen zero points\\
\hspace*{1em}if fixed budget: continue\\
\hspace*{1em}e = mean absolute error\\
\hspace*{1em}if e < best: best = e\\
\hspace*{1em}else: stop or freeze request
\end{quote}
The limit is $k$ for fixed updates and 20 otherwise. The masked request-local
loop executes all 20 steps, leaving frozen states unchanged. This is distinct
from compacting active requests or adding a host check to exit the full loop.
The upstream source also provides a no-early-stop optimizer; fixed iteration
counts are a simple existing strategy, not a new quantization algorithm.
That upstream variant uses different precision and step defaults, whereas
our fixed-count control preserves the legacy step.

\paragraph{Stopping precision and reconstruction magnitude.}
\label{app:stop-statistics}
We record every local update through the 20-update cap at initial compression
for all 384 original test pairs, including unchanged answers. Initial
raw target tensors and first logits agree across companions. All native
initial reconstructions exactly match their original recorded fingerprints.

The scalar precision ablation replaces
\texttt{abs(W-Wr).mean().float()} with
\texttt{abs(W-Wr).mean(dtype=float32)}. Subtraction, absolute values, local
FP16 updates, initialization, packing, BF16 metadata conversion and the final
update retained after a failed test are unchanged. Thus we change the stopping
statistic, not the quantization update precision. Shared and per-request
FP16/FP32 counts are read from the same 20-step local trajectories.
Target per-request error trajectories and counts match across companions in
all 26,112 initial comparisons, in both scalar precisions. This is an
initial-call diagnostic, not a new request-local generation comparison.

Across the two companion conditions, 32,873 of 52,224 native initial calls
fail the strict improvement test; 32,870 stop on equal FP16 errors. At 32,851
of those ties, the FP32 mean still improves over its earlier minimum.
The FP32 rule continues beyond the native endpoint in 28,798 of the tied
calls. All 4,053 improving ties that do not continue have already reached
the 20-update cap under both rules; they are not early terminations.
Thus scalar resolution often truncates an improving recorded trajectory;
higher precision still uses a shared stopping decision.

Full FP32 generation covers all 256 original test arithmetic pairs, with
native answers taken from the unchanged core records. The earlier 32
input-selected arithmetic pairs are repeated exactly and counted once.
QA retains the earlier 32 pairs (16 questions per model), selected by input-ID
hash before ablation generation. Original prompts, companions, batch size two,
precision, residual capacity and 512/64-token output caps are retained.
The complete arithmetic run adds 512 target answers and records actual cache
norms during initial and rolling compression. Stock-native, FP32-repeat and
BF16 controls pass on the longest input per model; all 64 repeated arithmetic
answers, their traces and cache fingerprints exactly match the earlier subset.
The prior 128 native subset answers also match the immutable original study.
Initial counts agree with those used in generation. These diagnostics reuse
test inputs; they are not an independent accuracy evaluation.
\begin{table*}[t]
\centering\small
\begin{tabular}{lrrrrrrrr}
\toprule
 & Initial pairs & Initial calls & \multicolumn{2}{c}{Different counts} & \multicolumn{2}{c}{Different caches} & \multicolumn{2}{c}{Changed answers / generated pairs} \\
Model/task & & & FP16 & FP32 & FP16 & FP32 & FP16 & FP32 \\
\midrule
Llama/math & 128 & 8,192 & 3,900 & 1,872 & 3,900 & 272 & 80/128 & 62/128 \\
Llama/QA & 64 & 4,096 & 1,473 & 316 & 1,473 & 160 & 0/16 & 0/16 \\
Qwen/math & 128 & 9,216 & 3,430 & 806 & 3,430 & 101 & 89/128 & 67/128 \\
Qwen/QA & 64 & 4,608 & 1,740 & 147 & 1,740 & 81 & 0/16 & 0/16 \\
\bottomrule
\end{tabular}
\caption{Decision precision with unchanged FP16 local updates. Initial calls cover all 384 test pairs (128 math and 64 QA per model), including unchanged answers. Each call pairs the same layer and key/value tensor across companions. Full arithmetic answers cover all 128 questions per model; QA retains the earlier input-selected 16-question subset per model. FP32 denotes the shared stopping mean, not model or update precision.}
\label{tab:precision}
\end{table*}

For a matched raw target $X$ and its reconstructed caches $\widehat X_A$ and
$\widehat X_B$, we measure the companion difference
$d=\|\widehat X_A-\widehat X_B\|_2/\|X\|_2$ and each condition's quantization
error $e_w=\|\widehat X_w-X\|_2/\|X\|_2$. Norms use actual reconstructed BF16
values, converted to FP32, including padded positions; they are not inferred
from hashes. Table~\ref{tab:precision-norms} separates the difference between
conditions from their error relative to the uncompressed target.
The median $d$ over all calls is zero in every stratum, so we report nonzero
medians with their denominators in Table~\ref{tab:precision}.
\begin{table}[t]
\centering\small
\begin{tabular}{lrrrr}
\toprule
 & \multicolumn{2}{c}{$d>0$ median (\%)} & \multicolumn{2}{c}{$e_w$ median (\%)} \\
Model/task & FP16 & FP32 & FP16 & FP32 \\
\midrule
Llama/math & 0.687 & 0.108 & 10.34 & 10.34 \\
Llama/QA & 0.576 & 0.035 & 10.48 & 10.48 \\
Qwen/math & 1.084 & 0.167 & 11.45 & 11.43 \\
Qwen/QA & 1.018 & 0.036 & 11.02 & 11.02 \\
\bottomrule
\end{tabular}
\caption{Actual initial-cache relative $L_2$ differences. The companion difference $d$ is summarized over nonzero calls only (counts in Table~\ref{tab:precision}); quantization error $e_w$ uses both conditions of every call. Calls and models sharing questions are not independent samples. These norms do not measure answer accuracy.}
\label{tab:precision-norms}
\end{table}

A large reduction in differing calls does not mean most requests become
unperturbed. With FP32, 189/256 arithmetic pairs still have at least one
different initial cache; 106 of these change answers. Another 23/67 pairs
with initial cache agreement change answers after rolling recompression
(Table~\ref{tab:precision-pairs}). Before the first output-token difference,
174 rolling calls have identical raw target inputs but different update counts
and reconstructed caches. We distinguish these from cache differences after
output histories diverge and restrict comparisons to both target lifetimes.
\begin{table}[t]
\centering\small
\begin{tabular}{lrr}
\toprule
Different initial caches & Llama & Qwen \\
\midrule
0 & 1/13 & 22/54 \\
1--4 & 57/107 & 45/74 \\
5--16 & 4/8 & -- \\
\bottomrule
\end{tabular}
\caption{FP32 shared stopping in all 256 test math pairs: changed full answers / pairs, grouped by the number of initial cache calls with different target reconstructions. A dash denotes an empty group. Bands are descriptive; initial cache agreement does not exclude later recompression differences.}
\label{tab:precision-pairs}
\end{table}

The initially agreeing cohorts differ between models: 13 Llama and 54 Qwen
questions remain, with median answer lengths of 170 and 251.5 tokens.
Later recompression occurs in 12/13 and 54/54 pairs. Before any output
difference, equal raw inputs yield different reconstructed caches in 12/13
and 44/54 pairs, yet only 1/13 and 22/54 change answers. Thus Llama also
experiences later perturbations without usually changing tokens. These unequal cohorts, conditioned on initial agreement, and different
models do not isolate why a cache
perturbation changes an answer; longer generation is descriptive exposure,
not an established explanation of the model gap.

Overall, 95 arithmetic pairs change answers under both scalar precisions,
74 only under native FP16, 34 only under FP32, and 53 under neither. The net
reduction from 169 to 129 is not a monotonic improvement for every question.
Confirmed numerical-correctness changes are 8/246 resolved pairs under native
and 7/248 under FP32. Neither these counts nor reconstruction norms establish
an answer-quality or latency advantage.

For illustration, Qwen's GSM8K test item 147 has identical initial caches
under FP32. After 128 generated tokens, the second layer's value-cache update
uses 20 versus 17 iterations on identical target inputs, giving a relative
reconstruction difference of $2.44\times10^{-4}$. The first token difference
appears at position 234; both final numbers remain correct. This example is
selected after analysis by the first input-ID hash among the 23 such pairs,
not used to estimate their frequency.

Both models use the original cache-study runtime on RTX A6000. The earlier
24-pair Qwen prefix from a shared device matches the complete initial-call
rerun and remains excluded from its 384-pair totals.

\paragraph{What additional updates improve.}
Table~\ref{tab:endpoints} measures actual initial-cache reconstructions after
0, 2, 4, 8 and 20 legacy updates on the earlier 64 input-selected pairs.
Here zero retains legacy FP16 initialization and differs from disabling
optimization. Fixed 20 has lower relative $L_2$ and mean absolute error than
fixed 4 in all 4,352 matched target calls. For example, median relative $L_2$
error falls from 10.416\% to 10.337\% for Llama arithmetic and from 11.724\%
to 11.440\% for Qwen arithmetic. The improvement is modest and does not imply
better answers. This supports retaining the original 20-update budget when
removing the shared early-stop test without tuning a smaller budget.
\begin{table*}[t]
\centering\small
\begin{tabular}{llrrrrr}
\toprule
Model/task & Error & $k=0$ & $k=2$ & $k=4$ & $k=8$ & $k=20$ \\
\midrule
Llama/math & $L_2$ (\%) & 10.7628 & 10.5295 & 10.4160 & 10.3455 & 10.3371 \\
Llama/math & MAE ($10^{-3}$) & 55.1617 & 54.0878 & 53.5362 & 53.0082 & 52.7284 \\
Llama/QA & $L_2$ (\%) & 10.9478 & 10.7150 & 10.5895 & 10.5285 & 10.5193 \\
Llama/QA & MAE ($10^{-3}$) & 57.6475 & 56.5702 & 55.9558 & 55.4022 & 55.1108 \\
Qwen/math & $L_2$ (\%) & 12.0867 & 11.8574 & 11.7241 & 11.6003 & 11.4400 \\
Qwen/math & MAE ($10^{-3}$) & 188.9711 & 184.7405 & 181.8135 & 178.8776 & 176.2468 \\
Qwen/QA & $L_2$ (\%) & 11.5697 & 11.3644 & 11.2323 & 11.0925 & 11.0026 \\
Qwen/QA & MAE ($10^{-3}$) & 185.7925 & 182.8034 & 180.7499 & 178.2636 & 175.9828 \\
\bottomrule
\end{tabular}
\caption{Median error of actual stored BF16 initial-cache reconstructions against the raw target. Each stratum uses 16 input-selected question pairs (1,024 Llama or 1,152 Qwen matched calls). Fixed-budget reconstructions agree across companions; shared questions/layers are not independent samples. $L_2$ is divided by the raw target norm. $k=0$ means legacy FP16 initialization with no updates, not the optimize=False baseline. These reconstruction errors do not measure answer quality.}
\label{tab:endpoints}
\end{table*}

\paragraph{Padding with fixed attended values.}
We hold BF16 attended tensors of shape $[2,8,128,128]$ fixed for 16 seeds,
prepend 0, 32, 128 or 512 padding positions, and retain axis-1 groups of 64.
Padding is either zero or a repeated per-head vector, generated once per seed.
We compare each nonzero length with its unpadded counterpart on the target's
attended reconstruction: 96 comparisons per method from 16 tensors, two
patterns and three lengths. Request-local stopping changes the target count
and attended reconstruction in 58/96 comparisons (24/48 zero-padding;
34/48 repeated-vector), whereas fixed 20 changes neither in any comparison.
Equal target counts always give equal attended reconstructions. This is a
tensor-level mechanism check: these synthetic padding values and correlated
comparisons do not estimate a serving failure rate. It adds no padding-aware
repair or answer-quality claim.

\section{Test-Set Quality and Coverage}
\label{app:quality}
The test-set seven-method comparison has 128 arithmetic and 64 QA targets
per model with two companions. Official arithmetic parsing, explicit final
numbers, QA F1, and exact match remain separate. We average the two companions
within a target for quality comparisons; they are not independent samples.
Table~\ref{tab:prospective-quality} reports all methods. The numerical bounds
assign unresolved assertions either zero or one correctness, without claiming
that a valid final number verifies the reasoning.
Unresolved answers lack a unique explicit final number; they are not known
errors (Appendix~\ref{app:scoring}). Table~\ref{tab:compression} separately
reports tensor-compression times, which are not generation latency.

Request-local stopping preserves termination on the same padded request tensor,
but the data
do not establish that this is more accurate than a fixed count.
On test-set Llama arithmetic, its final-number accuracy is bounded by
84.38--89.06\%, versus 86.72--90.62\% for fixed 4 and 87.50--91.41\%
for fixed 20. These bounds are not confidence intervals. Neither native nor
repaired HQQ consistently improves over BF16 across models and tasks.
The paired comparisons below include Qwen and Quanto.

\begin{table*}[t]
\centering
\small
\begin{tabular}{lrllrl}
\toprule
Model & GSM8K test ID & Assertion A & Assertion B & Reference & A to B \\
\midrule
Llama & 439 & \texttt{\#\#\#\# 17} & \texttt{\#\#\#\# 7} & 17 & correct $\to$ wrong \\
Llama & 1310 & \texttt{\#\#\#\# 64} & \texttt{\#\#\#\# 32} & 64 & correct $\to$ wrong \\
Llama & 1092 & \texttt{\#\#\#\# 2160} & \texttt{\#\#\#\# 1080} & 1080 & wrong $\to$ correct \\
Llama & 406 & \texttt{\#\#\#\# 200} & \texttt{\#\#\#\# 120} & 200 & correct $\to$ wrong \\
Llama & 532 & \texttt{\#\#\#\# 25} & \texttt{\#\#\#\# 35} & 25 & correct $\to$ wrong \\
Llama & 648 & \texttt{\#\#\#\# 56} & \texttt{\#\#\#\# 54} & 54 & wrong $\to$ correct \\
Qwen & 769 & \texttt{\#\#\#\# 365} & $\boxed{110}$ & 110 & wrong $\to$ correct \\
Qwen & 373 & $\boxed{240000}$ & $\boxed{320000}$ & 240000 & correct $\to$ wrong \\
\bottomrule
\end{tabular}
\caption{All eight confirmed cache-test numerical correctness changes. Assertions display the recognized marker or box and normalized number, not the entire answer or its exact currency formatting. Full raw outputs and extraction anchors are included in the artifact. Companion labels A and B are arbitrary; five changes lose correctness and three gain it.}
\label{tab:correctness-cases}
\end{table*}
Table~\ref{tab:correctness-cases} gives all eight confirmed changes under the
fixed companion ordering A to B: Llama has four correct-to-wrong and two
wrong-to-correct pairs; Qwen has one in each direction. A and B have no
intrinsic quality ordering. Complete questions, both outputs and extractor
anchors are supplied, including the two Qwen cases where boxes or currency
formatting differ. These are final-number checks, not reasoning judgments.

\begin{table}[htbp]
\centering
\small
\begin{tabular}{lrrr}
\toprule
Pass changes & Same & Different & Unresolved \\
\midrule
Yes & 5 & 6 & 5 \\
No & 232 & 3 & 5 \\
\bottomrule
\end{tabular}
\caption{Official pass changes versus final numerical values in all 256 test-set native-HQQ arithmetic pairs. Same and Different require resolved numbers on both sides; Unresolved means at least one extraction fails. All 16 official changes occur in this arithmetic subset.}
\label{tab:pass-value-transitions}
\end{table}

The official math judgment changes in 9/128 Llama pairs and 7/128 Qwen pairs,
totaling 16/256; neither model changes the official QA judgment.
This criterion requires the correct number after the prescribed
\texttt{\#\#\#\#} marker: \texttt{Answer: 6} fails even when 6 is correct.
The separate final-number extractor does not require that marker.

Table~\ref{tab:pass-value-transitions} connects the two numerical endpoints.
Of nine pairs with changed final values, six change both official pass and
final-number correctness. Two change final-number correctness while official
pass stays unchanged, and one changes between two incorrect values with
unchanged official pass. Five additional official changes retain the same
resolved number, and five involve an unresolved side. These categories
explain all 16 official changes without treating extraction failure as a
known numerical error.
\begin{table*}[tp]
\centering
\small
\begin{tabular}{lrrrrrr}
\toprule
& \multicolumn{3}{c}{Llama} & \multicolumn{3}{c}{Qwen} \\ 
Method & GSM strict & Final value & Hotpot F1 & GSM strict & Final value & Hotpot F1 \\
\midrule
BF16 & 78.12 & 89.06--90.62 & 70.71 & 66.41 & 90.62--96.09 & 65.08 \\
Native & 74.61 & 87.11--90.23 & 69.82 & 69.14 & 93.36--95.70 & 62.69 \\
Unoptimized & 71.88 & 83.59--87.50 & 69.82 & 64.84 & 91.41--95.31 & 65.02 \\
Fixed 4 & 75.00 & 86.72--90.62 & 69.82 & 66.41 & 92.97--96.09 & 65.56 \\
Fixed 20 & 75.00 & 87.50--91.41 & 69.82 & 67.19 & 93.75--96.09 & 62.69 \\
Request-local & 71.09 & 84.38--89.06 & 69.82 & 70.31 & 92.97--95.31 & 62.67 \\
Quanto & 75.00 & 85.16--89.06 & 69.82 & 67.19 & 94.53--97.66 & 65.24 \\
\bottomrule
\end{tabular}
\caption{Test-set quality in percent, averaging the two companion conditions within each target: 128 GSM8K and 64 HotpotQA questions. Final value reports lower and upper correctness bounds obtained by treating unresolved explicit numerical assertions as incorrect or correct, respectively; it is not a reasoning score. Methods other than BF16 use four-bit caches. Fixed four updates was included after inspecting development results, before test-set generation; it is not a proven optimal budget.}
\label{tab:prospective-quality}
\end{table*}
Llama's six compressed methods share HotpotQA F1 69.82 because every paired
answer has the same F1 as native HQQ, not because all outputs are identical.
Against native, no optimization changes six of 128 answer strings, fixed 4
and request-local change two each, and Quanto changes four; fixed 20 changes
none. The individual token and text records distinguish these cases.
\begin{table*}[tp]
\centering
\small
\begin{tabular}{llrrrr}
\toprule
Dataset & Model & $N$ & Text $\Delta$ & Pass $\Delta$ & Correct $\Delta$/all \\
\midrule
GSM8K & Llama & 128 & 62.50 [53.91, 71.09] & 7.03 [3.12, 11.72] & 4.69 [1.56, 8.59] \\
GSM8K & Qwen & 128 & 69.53 [61.72, 77.34] & 5.47 [2.34, 10.16] & 1.56 [0.00, 3.91] \\
HotpotQA & Llama & 64 & 0.00 [0.00, 0.00] & 0.00 [0.00, 0.00] & -- \\
HotpotQA & Qwen & 64 & 1.56 [0.00, 4.69] & 0.00 [0.00, 0.00] & -- \\
\bottomrule
\end{tabular}
\caption{Observed native four-bit change percentages and descriptive 95\% paired question-cluster bootstrap intervals. Every denominator is all target questions in that row. Resolved correctness changes count only pairs whose final numerical assertions resolve on both sides; unresolved pairs may contain additional changes and are not established negatives. Models share questions. Intervals are unadjusted, and a zero-event interval is not a population upper bound.}
\label{tab:prospective-rate-intervals}
\end{table*}

Table~\ref{tab:prospective-rate-intervals} reports companion-dependent change
rates among all targets. Unlike Table~\ref{tab:prospective}, its correctness-change
denominator includes unresolved pairs. Within each model and task, descriptive
95\% percentile intervals use 2,000 bootstrap resamples of target questions,
keeping both companion conditions together. They are unadjusted for multiple
comparisons and do not account for all semantic dependence between questions.
A zero-event interval is not a population risk bound.
The official scorer is unchanged; the explicit-number extractor's patterns
are reference-blind and frozen before collection.

\section{Complete Compression Cost}
\label{app:cost}
\paragraph{Actual initial KV tensors.}
Table~\ref{tab:actual-cost} measures complete compression on model-produced
inputs. Before collection, we select two test questions per model/task by
hashing input IDs, without inspecting answers. For both companions we capture
keys and values from the first, middle and last layers: eight model/question
pairs yield 96 batch tensors. All initial target logits and raw/reconstructed
cache hashes match the original diagnostic. Native counts and fixed-20
reconstruction checks are collected outside timing; timed paths have no
tracing. Two warmups precede nine synchronized wall-time measurements with
shuffled method order. No foreign process was observed on either timing GPU;
the two models use separate devices, with shared CPU and bus resources.
\begin{table*}[t]
\centering\small
\begin{tabular}{lrrrrrr}
\toprule
 & \multicolumn{3}{c}{Native update count} & \multicolumn{3}{c}{Complete compression (ms)} \\
Model/task & Median & Range & Below 20 & Native & Fixed 20 & No optimization \\
\midrule
Llama/math & 13 & 9--20 & 20/24 & 4.46 & 4.11 & 0.48 \\
Llama/QA & 14 & 10--20 & 16/24 & 7.58 & 8.75 & 0.38 \\
Qwen/math & 19.5 & 9--20 & 12/24 & 3.93 & 3.56 & 0.34 \\
Qwen/QA & 20 & 10--20 & 11/24 & 12.25 & 11.69 & 1.18 \\
\midrule
All cells & 18 & 9--20 & 59/96 & 6.35 & 6.22 & 0.52 \\
\bottomrule
\end{tabular}
\caption{Actual initial-KV compression on RTX A6000. Each stratum uses two input-selected questions, two companions, three layers and keys/values: 24 batch tensors. Times are geometric means of per-tensor medians from nine repetitions after two warmups. Count tracing and correctness checks are outside timing. Initial-cache compression excludes rolling updates and answer generation; reused inputs and layers are not independent samples.}
\label{tab:actual-cost}
\end{table*}

Native counts range from 9 to 20, with 59/96 below the cap. Fixed 20 is
15.4\% slower for Llama QA but faster in the other three strata. No
optimization is substantially cheaper. This small initial-cache sample does
not establish a general speed ranking, rolling-compression cost or generation
latency. Nor does it establish that optimization is necessary for answer
quality: its reconstruction objective and answer correctness are distinct.
All 2,592 timings, input bindings and the full count histogram are supplied.

\paragraph{Synthetic tensor conditions.}
Table~\ref{tab:compression} compares complete four-bit compression on 36
conditions. The design crosses batch sizes 2/4/8, eight heads, head dimension
128, lengths 128/512/2048, BF16/FP16, and two synthetic distributions. Normal
inputs are standard Gaussian samples. Heterogeneous inputs scale successive
requests cyclically by 0.125/0.5/2/8, then multiply every 257th flattened entry,
starting at zero, by four, before casting to the input dtype. Every timing
input matches its hash in the frozen condition manifest. Two warmups precede
nine synchronized measurements per method and condition, with fixed method-order
seeds. No other GPU compute process was observed on the RTX A6000 timing device;
the CPU remained shared.
\begin{table}[htbp]
\centering
\small
\begin{tabular}{lr}
\toprule
Method & Time (ms) \\
\midrule
Native & 8.3345 \\
Unoptimized & 1.2849 \\
Fixed 2 & 2.5221 \\
Fixed 4 & 3.1205 \\
Fixed 8 & 4.3243 \\
Fixed 20 & 7.8219 \\
Request-local & 12.9050 \\
Quanto & 0.4933 \\
\bottomrule
\end{tabular}
\caption{Complete 4-bit tensor compression on 36 matched conditions: initialization, optimization, packing, and metadata conversion. Times are geometric means of per-condition medians over nine repetitions, in milliseconds; two warmups are excluded. All methods store 0.28125 times the original tensor bytes.}
\label{tab:compression}
\end{table}

Time includes initialization, optimization, packing, and metadata conversion;
input loading is excluded. Request-local stopping uses the same implementation
as the test-set quality/control study: the full 20-update loop runs on all
rows, while device masks freeze each request's final state without reading
stopping flags on the host. Fixed-20 timing excludes unused diagnostic
initialization. Matched controls verified identical compressed outputs and
reconstructions to the quality evaluation implementation; the two implementations
need not have equal latency.

Geometric means of condition medians are \LocalMs{} ms for request-local
stopping, \NativeMs{} ms for native HQQ, \FixedFourMs{} ms for fixed 4, and
\FixedTwentyMs{} ms for fixed 20. Quanto is faster in this benchmark.
Across conditions, the median within-condition absolute deviation divided
by the timing median is 0.76\% for native, 1.50\% for fixed 4, 0.79\% for
fixed 20 and 0.45\% for request-local. Raw repetitions are supplied.
Native performs a scalar continuation check at each iteration, whereas the
fixed loop omits it; thus more updates need not imply more time. The relative
cost of that check is not separately measured. Synthetic distributions do
not establish real-cache iteration counts or serving latency.
A separate optimizer-only benchmark also evaluates early exit by compacting
active requests. Across its 36 four-bit conditions, this costs 6.0038 ms,
versus 9.4352 ms for the masked loop and 5.9445 ms for fixed 20. The artifact
includes all 72 two-/four-bit conditions and periodic host-check variants.
These exclude initialization and packing, so they cannot be compared directly
with the complete-compression times above or establish generation speed.

All methods store 28.125\% of the original tensor payload, including codes
and reconstruction metadata. This excludes model weights and temporary
workspace; these measurements establish neither total-model memory savings
nor end-to-end generation speed.

\paragraph{Boundary under natural rebatching.}
\label{app:rebatch-tasks}
\begin{table}[t]
\centering\small
\begin{tabular}{lrrrr}
\toprule
 & \multicolumn{2}{c}{Shuffled} & \multicolumn{2}{c}{Sorted} \\
Method & Long & Short & Long & Short \\
\midrule
BF16 & 74 & 1 & 82 & 3 \\
Native HQQ & 115 & 5 & 111 & 11 \\
No optimization & 97 & 6 & 93 & 8 \\
Fixed 2 & 92 & 5 & 93 & 10 \\
Fixed 4 & 94 & 7 & 98 & 11 \\
Fixed 8 & 102 & 4 & 106 & 11 \\
Fixed 20 & 99 & 6 & 102 & 14 \\
Request-local & 98 & 5 & 103 & 13 \\
Quanto & 94 & 5 & 98 & 12 \\
\bottomrule
\end{tabular}
\caption{Changed development answers after rebatching, relative to source order. Long: 128 math/instruction pairs; short: 256 QA pairs. Batches contain four questions. All quantized methods use four bits; padding and numerical execution may change.}
\label{tab:rebatch}
\end{table}
Table~\ref{tab:rebatch} reports both rearrangements of the development set.
Each method's per-model, per-task pairing counts, first differing token
positions, score changes, and all 10,368 outputs are supplied. Ordinary padding
and default attention allow numerical effects in addition to stopping changes.
For shuffled layouts, native/fixed-4 discordant counts are 25 native-only and
six fixed-4-only changes; for length sorting they are 21 and eight.
For native/request-local the corresponding counts are 23/six and 16/ten.
Questions are linked by batch membership and shared source documents; each
model/task forms one connected component. Treating all 384 comparisons as
independent McNemar trials would understate dependence. Each development
layout has only one realization; the separate test-set replication below
uses six shuffle seeds.

Table~\ref{tab:rebatch} separates long-answer arithmetic and instruction
following from short QA for every method. Native changes 115/128 long-answer
texts after shuffling and 111/128 after sorting, compared with 74/128 and
82/128 for BF16, and 99/128 and 102/128 for fixed 20. Short-answer changes
are not uniformly lower with repairs: sorted fixed 20 changes 14/256, versus
11/256 for native. These task categories follow prompts, not observed lengths.
Table~\ref{tab:rebatch-correctness} compares all methods' arithmetic correctness
changes. The native/fixed-20 counts are three/one under each rearrangement,
with different unresolved coverage. String changes greatly outnumber confirmed
correctness changes; neither metric establishes an average accuracy loss.
First-token differences also occur, which the controlled study excludes.
\begin{table}[t]
\centering\small
\begin{tabular}{lrrrr}
\toprule
 & \multicolumn{2}{c}{Shuffled} & \multicolumn{2}{c}{Sorted} \\
Method & Strict & Num. & Strict & Num. \\
\midrule
BF16 & 2 & 2/61 & 6 & 2/62 \\
Native HQQ & 8 & 3/61 & 3 & 3/62 \\
No optimization & 7 & 1/60 & 8 & 1/59 \\
Fixed 2 & 5 & 1/60 & 8 & 4/60 \\
Fixed 4 & 7 & 1/60 & 10 & 2/63 \\
Fixed 8 & 8 & 2/61 & 5 & 2/62 \\
Fixed 20 & 4 & 1/60 & 8 & 1/60 \\
Request-local & 5 & 2/60 & 4 & 1/62 \\
Quanto & 7 & 3/60 & 5 & 3/61 \\
\bottomrule
\end{tabular}
\caption{Development GSM8K correctness changes under natural rebatching. Strict: changed strict-marker judgment out of 64 pairs. Num.: changed final-number correctness / pairs resolved in both answers. Unresolved pairs are excluded from that denominator, not treated as errors.}
\label{tab:rebatch-correctness}
\end{table}

\paragraph{Test-set replication across shuffle seeds.}
We reuse the frozen 128 arithmetic and 64 QA test questions per model,
batching four questions within each task with natural left padding and
stock scaled dot-product attention (SDPA). BF16, native HQQ and fixed 20 run source order, length sorting
and six hash-derived shuffles: 9,216 answers. Prompts, model revisions,
BF16 weights, HQQ groups and the 512/64-token caps are unchanged. The inputs,
seeds, methods and primary contrast are fixed before these runs. Fixed 20
retains the original maximum budget without fitting a smaller count.
Every question appears once per model, method and layout. Each worker's
longest-input controls require repeated outputs and first-logit agreement
across methods; common controls also agree across workers.
\begin{table*}[t]
\centering\small
\begin{tabular}{lrrrrrr}
\toprule
 & \multicolumn{3}{c}{Changed text / 384 pairs} & \multicolumn{3}{c}{Math correctness changes / resolved pairs} \\
Shuffle seed & BF16 & Native & Fixed 20 & BF16 & Native & Fixed 20 \\
\midrule
1 & 136 & 218 & 168 & 3/248 & 12/248 & 9/249 \\
2 & 130 & 218 & 184 & 3/248 & 12/250 & 9/247 \\
3 & 144 & 215 & 190 & 3/247 & 10/251 & 11/249 \\
4 & 141 & 205 & 183 & 3/247 & 11/252 & 9/250 \\
5 & 138 & 197 & 175 & 3/248 & 10/249 & 11/251 \\
6 & 142 & 213 & 173 & 2/246 & 9/251 & 13/250 \\
\midrule
Length-sorted & 147 & 213 & 194 & 5/246 & 12/250 & 11/250 \\
\bottomrule
\end{tabular}
\caption{Natural rebatching of the frozen test set: two models and 192 questions, compared with each method's source order. Six shuffle seeds are the paired units of the sole primary descriptive contrast; length sorting is outside that test. Numerical correctness uses the unchanged final-number extractor. Padding and kernels may change, so these differences do not isolate stopping.}
\label{tab:test-rebatch}
\end{table*}

Each method is compared with its own source-order answers. The sole primary
contrast is the seed-level native-minus-fixed-20 text-change fraction, pooling
the two models and all 192 questions. Its mean is 8.38 percentage points
(range 5.73 to 13.02); a two-sided enumeration of all 64 sign assignments
gives $p=0.03125$ under symmetric seed-level differences. This is conditional
on the fixed questions and models, not a question-independent test or an
isolated causal effect of stopping. Length sorting, task/quality breakdowns
and all other metrics are descriptive. Source questions are reused across
seeds; this is a new batching comparison, not an untouched accuracy holdout.

Table~\ref{tab:test-rebatch} also reports confirmed arithmetic correctness
changes, retaining unresolved pairs. Table~\ref{tab:test-quality} separates text changes by model/task and compares
answer quality over all eight layouts. These repetitions do not establish
quality superiority. Per-model/task changes, output lengths, capped answers,
first-token differences and all individual results accompany the artifact.
\begin{table*}[t]
\centering\small
\begin{tabular}{llrrr}
\toprule
Model/task & Measure & BF16 & Native & Fixed 20 \\
\midrule
Llama/math & Text $\Delta$ & 50.91 & 79.69 & 69.27 \\
 & Numeric bounds & 88.18--88.87 & 87.79--89.65 & 87.21--89.45 \\
Llama/QA & Text $\Delta$ & 1.30 & 5.47 & 4.69 \\
 & EM/F1 & 55.47/70.24 & 52.54/68.94 & 52.34/69.08 \\
Qwen/math & Text $\Delta$ & 54.95 & 79.69 & 65.62 \\
 & Numeric bounds & 90.92--96.48 & 93.65--96.09 & 94.24--96.09 \\
Qwen/QA & Text $\Delta$ & 3.39 & 5.47 & 4.95 \\
 & EM/F1 & 48.44/64.89 & 47.85/63.50 & 48.24/63.68 \\
\bottomrule
\end{tabular}
\caption{Test rebatching by model and task (\%). Text $\Delta$ averages changes over six shuffles, with 128 math or 64 QA pairs per model. Quality averages the same questions over all eight layouts. Numeric bounds treat unresolved answers first as wrong, then as correct; they are not confidence intervals. QA gives answer exact match / F1. Repetitions are not independent accuracy samples; no quality superiority is established.}
\label{tab:test-quality}
\end{table*}

\paragraph{Additional repair studies with negative outcomes.}
We also tested key/value-specific budgets and optimizing only appended cache
entries, with uniform budgets 0/2/4/8/20 and reference methods. This separate
study collected 6,912 answers across two models on 256 arithmetic and 128 QA
questions. The prespecified primary contrast was Qwen arithmetic correctness
with four key updates and no value updates versus no optimization. Its
2.73-point increase did not meet the 0.05 criterion ($p=0.05868$), and its
uncertainty bounds included losses when unresolved answers were retained.
The artifact retains all
40 contrasts, including the 39 descriptive secondary comparisons. It does
not establish a practical advantage for an adaptive repair.
A separate 864-call implementation study compares six execution paths under
natural and fixed-token replay generation, including warmups. Its 24 plans
reuse 48 questions across two models, with only two measured calls per
plan/method/mode. Neither study supports a general quality or latency ranking;
their recorded paired endpoints, timings and analysis procedures are supplied.

\section{Weight-Compression Invocation Partition}
\label{app:weights}
TorchAO's released Llama example can concatenate query/key/value projection
weights before applying HQQ \citep{torchao2026qkv}. Its inspected HQQ
weight-only path invokes the same quantizer \citep{torchao2026hqq}.
We compare that function on concatenated or separate Q/K/V matrices, preserving
their values and every 64-element group. Concatenation is along output rows
with axis-1 quantization, so groups are not mixed. The relevant quantization
functions are unchanged between the inspected revision and version 0.13.
This is source and numerical-function evidence, not a complete
TorchAO deployment test. Some public kernels instead concatenate already
quantized packets; they are not counted as affected paths.

For whole-model evaluation, we quantize the seven Transformer linear weights
per block, retaining embeddings, normalization, and output-head weights.
We reconstruct every method with the same FP32 affine expression followed
by a BF16 cast and run the same unfused BF16 model and cache. Neither int4
kernel speed nor physical memory savings are measured.

\paragraph{Three-model weight-compression evaluation.}
We sample 256 GSM8K test and 128 HotpotQA validation questions before generation,
excluding all prior source IDs and normalized-question hashes found in the
recorded exposure scan. Hash ranking with seed 922451 selects from 679 eligible
arithmetic and 4,790 eligible QA questions. These exclusions are at the question
level, not the document or pretraining level. All three models answer the same
384 questions: Llama-3.1-8B-Instruct, Qwen3-8B with thinking disabled, and
Mistral-7B-Instruct-v0.3, at the revisions in Appendix~\ref{app:settings}.
The 1,152 native model--question comparisons are not independent questions.

Each model runs BF16, native separate, native joint, no optimization, and fixed
2/4/8/20, giving 9,216 answers in 2,304 four-question generation calls. The
paired native methods use identical tokenized batches, prompts, greedy decoding,
and output caps of 512 arithmetic or 64 QA tokens. No input exceeds the
8,192-token cap; no input is truncated. All quantized methods use four bits,
group size 64, the unchanged HQQ update ordering and FP16 fitting defaults,
and identical BF16 reconstructed-weight inference with an unquantized cache.
The fixed method order, generated with seed 922452, is BF16, native separate,
native joint, fixed 8, 2, 4, 20, and no optimization.

Generation uses three RTX A6000 GPUs, PyTorch 2.14.0+cu126, Transformers 5.17.0,
and scaled dot-product attention. The later weight study uses a separately
pinned runtime; the completed cache study retains PyTorch 2.7.0. Each paired
comparison uses one runtime, and results are not a cross-version comparison. This study measures answer dependence,
not deployment speed or memory savings. Generation resumed after an
interruption only after matching original and compressed weight hashes and
runtime metadata and reproducing a continuity check. Previously completed
outputs were unchanged, and duplicate control outputs were excluded.

\paragraph{Results and controls.}
\begin{table}[htbp]
\centering
\small
\begin{tabular}{lrrr}
\toprule
Numerical outcome & Llama & Qwen & Mistral \\
\midrule
Same correct & 202 & 224 & 29 \\
Correct $\to$ wrong & 4 & 2 & 2 \\
Wrong $\to$ correct & 3 & 1 & 5 \\
Different wrong & 5 & 1 & 12 \\
Same wrong & 22 & 13 & 8 \\
Unresolved & 20 & 15 & 200 \\
\bottomrule
\end{tabular}
\caption{Final-number outcomes under four-bit HQQ weight compression. Direction is separate to joint; each column partitions all 256 math pairs. Unresolved pairs have at least one unextracted final number. Text-change counts appear in the accompanying results.}
\label{tab:weight-fresh}
\end{table}
Joint versus separate compression changes 55/96, 77/108, and 64/96 reconstructed
Q/K/V matrices for Llama, Qwen, and Mistral, respectively. Non-QKV weights agree.
All 300 matrix-level bidirectional count interventions reproduce the opposite
packets. All 500 fixed-method layer checks (100 layers, five budgets including
zero) preserve partition invariance. These checks establish the numerical
compression mechanism; they are not another answer-level causal replay study.

Arithmetic texts change in 162/256, 184/256, and 208/256 pairs; QA texts change
in 7/128, 6/128, and 14/128. QA exact-match judgments change in zero, three,
and three pairs. Confirmed arithmetic correctness changes total seven, three,
and seven, with both correct-to-wrong and wrong-to-correct transitions
(Table~\ref{tab:weight-fresh} shows all numerical transitions). Token-sequence and text-change counts
coincide in this study. Larger text-change counts do not imply that every
changed answer is a content error.

The unchanged final-number extractor leaves 20, 15, and 200 arithmetic pairs
with an unresolved side. Mistral has 181 unresolved answers in each native
condition: 169 of these in separate compression and 165 in joint compression
reach EOS, so missing recognized final-number assertions are not mainly a
length-cap effect. In each condition, 166 of the 181 unresolved outputs
contain the requested answer marker but fail the extractor's accepted syntax.
For example, a marked number followed by ``miles per hour'' exceeds its
single-unit grammar; an empty marker supplies no number. Of the remaining
outputs, separate compression has 13 without recognized answer syntax and
two with unparsed answer clauses; joint compression has 15 without recognized
syntax. These automatic syntax categories explain extraction coverage without
assigning semantic correctness. The frozen extractor and scores are unchanged.
Unresolved answers are neither counted as wrong nor discarded. The coverage deficit prevents a general accuracy ranking; in
particular, Mistral's BF16 count of 21 verified correct answers with 218
unresolved is not an estimate that only 21 answers are correct.
Table~\ref{tab:weight-fresh-quality} reports all eight methods. The study does
not establish superiority of adaptive stopping or a universally best fixed
budget.

Descriptive 95\% percentile intervals use 10,000 bootstrap samples of whole
four-question batches, with seed 922453 and 64 arithmetic or 32 QA batches
per model. Resampling is separate by model and task; models sharing questions
are not pooled as independent samples. The artifact reports all nine registered endpoints and their intervals
for every model, without multiplicity adjustment.
Confirmed-correctness intervals describe the observed count only; they do not
resolve unknown answers. A zero-event interval is not a population upper bound.
\begin{table*}[htbp]
\centering
\small
\begin{tabular}{llrrrrr}
\toprule
Model & Method & Final math numbers & Math official & QA EM & QA F1 & Length limit \\
\midrule
Llama & BF16 & 221/28/7 & 174 & 55.47 & 71.02 & 5 \\
 & Native separate & 208/36/12 & 160 & 56.25 & 71.06 & 11 \\
 & Native joint & 208/35/13 & 161 & 56.25 & 71.06 & 10 \\
 & No optimization & 209/36/11 & 165 & 56.25 & 70.38 & 10 \\
 & Fixed 2 & 209/35/12 & 161 & 57.03 & 70.10 & 11 \\
 & Fixed 4 & 219/27/10 & 169 & 55.47 & 69.47 & 8 \\
 & Fixed 8 & 208/35/13 & 162 & 56.25 & 69.98 & 12 \\
 & Fixed 20 & 209/33/14 & 159 & 55.47 & 69.87 & 14 \\
\midrule
Qwen & BF16 & 238/10/8 & 143 & 62.50 & 75.73 & 3 \\
 & Native separate & 230/15/11 & 98 & 55.47 & 68.90 & 7 \\
 & Native joint & 227/16/13 & 91 & 54.69 & 68.60 & 8 \\
 & No optimization & 235/9/12 & 144 & 54.69 & 68.05 & 4 \\
 & Fixed 2 & 224/11/21 & 134 & 57.81 & 70.97 & 7 \\
 & Fixed 4 & 228/15/13 & 109 & 54.69 & 67.94 & 4 \\
 & Fixed 8 & 228/14/14 & 99 & 57.03 & 69.81 & 10 \\
 & Fixed 20 & 230/15/11 & 85 & 55.47 & 68.81 & 7 \\
\midrule
Mistral & BF16 & 21/17/218 & 7 & 33.59 & 55.48 & 12 \\
 & Native separate & 40/35/181 & 27 & 36.72 & 56.97 & 13 \\
 & Native joint & 43/32/181 & 28 & 37.50 & 57.55 & 18 \\
 & No optimization & 60/24/172 & 42 & 35.16 & 55.07 & 11 \\
 & Fixed 2 & 53/23/180 & 39 & 33.59 & 54.44 & 15 \\
 & Fixed 4 & 53/38/165 & 30 & 33.59 & 54.61 & 10 \\
 & Fixed 8 & 49/27/180 & 33 & 35.16 & 55.62 & 12 \\
 & Fixed 20 & 41/32/183 & 33 & 36.72 & 56.60 & 13 \\
\bottomrule
\end{tabular}
\caption{All eight methods in the three-model weight-compression evaluation. Each method answers 256 math and 128 QA questions per model. Final math numbers count correct/wrong/unresolved outcomes; official math counts passes under the strict marker parser. QA EM and F1 are percentages. Length limit counts capped outputs across both tasks. Identical BF16 inference on reconstructed weights and an unquantized cache isolates the numerical compression choices; it does not measure int4 deployment speed or memory.}
\label{tab:weight-fresh-quality}
\end{table*}

\section{Artifact and Verification Scope}
\label{app:artifact}
The central cache-study code and verification records are available at \ArtifactURL.
The accompanying source archive contains all 384 cache test-set pairs, with
complete output text and token IDs for seven methods, references, scores,
and count-replay records with cache fingerprints and controls. It also
contains all 9,216 answers from the three-model weight-compression evaluation,
including inputs, references, token IDs, scores, batch plans, and quantization
manifests. Exploratory task-analysis records accompany the verification scripts.

The portable audits check file integrity, allocation completeness, arithmetic
scores, recorded count transfer, and matrix controls. Original cache/weight QA
scores are supplied; the new test-rebatch QA answers are rescored with the
pinned answer EM/F1 functions. These audits operate on saved records; they do
not regenerate model outputs or hidden states. Cache fingerprints are not
complete tensors.

The archive supplies executable subset reproduction with pinned inputs,
model revisions, dependency instructions and explicit-mask/count-replay adapters;
model checkpoints are obtained separately. Historical source excerpts are
also retained for inspection. The archive additionally includes
all development and test natural-rebatching outputs, full arithmetic FP32
records, measured initial/rolling norms, fixed-budget error curves, padding
controls and raw complete-compression timing samples, with portable verification. Native two- and four-bit output records for all 960 five-model development
pairs are also included, allowing the 404/960 and 227/960 text-change totals
to be recomputed. Other development methods remain outside this extract. New GPU execution
uses the supplied prompts and requires licensed checkpoints and the recorded runtime. The archive documentation
provides verification commands and maps each result to its records.

\section{Exploratory Analysis of Task Differences}
\label{app:task-difference}
We conduct an exploratory post hoc analysis of the recorded native cache
and weight outputs to compare math and document-QA sensitivity.
Table~\ref{tab:task-difference} reports each model and task separately;
questions are shared across models and are not pooled as independent samples.
\begin{table*}[htbp]
\centering
\small
\begin{tabular}{lllrrrr}
\toprule
Study & Model & Task & Median length & Full changes & First 64 & After 64 only \\
\midrule
Cache & Llama & Math & 202 & 80/128 & 39/128 & 41/128 \\
Cache & Llama & QA & 5 & 0/64 & 0/64 & 0/64 \\
Cache & Qwen & Math & 266.5 & 89/128 & 36/128 & 53/128 \\
Cache & Qwen & QA & 5 & 1/64 & 1/64 & 0/64 \\
Weight & Llama & Math & 216 & 162/256 & 80/256 & 82/256 \\
Weight & Llama & QA & 5 & 7/128 & 7/128 & 0/128 \\
Weight & Mistral & Math & 249 & 208/256 & 114/256 & 94/256 \\
Weight & Mistral & QA & 7 & 14/128 & 14/128 & 0/128 \\
Weight & Qwen & Math & 278 & 184/256 & 99/256 & 85/256 \\
Weight & Qwen & QA & 4.5 & 6/128 & 6/128 & 0/128 \\
\bottomrule
\end{tabular}
\caption{Exploratory analysis of recorded native comparisons. Median lengths include EOS and combine both conditions. Full changes count differing token sequences; their counts equal text changes here. First 64 counts pairs already differing within that prefix; After 64 only counts later first differences. QA early termination is retained, so this is not a controlled length experiment.}
\label{tab:task-difference}
\end{table*}

QA prompts supply documents and request only the shortest answer without
explanation. Math prompts request reasoning followed by a final number.
Generated lengths include EOS, and medians combine both native conditions
within a model/task. For each pair we locate the first differing generated
token, including EOS. Comparing prefixes of 10, 32 and 64 tokens leaves early
EOS intact; it neither regenerates answers nor equalizes their realized
lengths. The accompanying records and analysis code supply all thresholds
and model/task summaries.

In the cache study, QA median length is 5 tokens for both models, while math
medians are 202 and 266.5. All QA outputs terminate by EOS before the 64-token
cap; their maximum lengths are 28 and 42. Of the 169 changed math pairs, 94
first diverge after token 64. A common 64-token comparison still gives 39/128
and 36/128 changed math pairs versus 0/64 and 1/64 QA pairs. Thus a larger
output cap alone does not establish the task difference's cause: QA usually
ends much earlier even inside that common window.

All 128 QA pairs have different initial reconstructed target-cache hashes,
with identical raw initial target inputs. In 127 of them the complete text
nevertheless stays unchanged. The low QA text-change count therefore does
not mean the compressor was unaffected. QA has one initial compression step
before the target ends and no later recompression. Math includes later
recompression in 240/256 Llama and 253/256 Qwen native answers. These counts
exclude compression events after the target has ended. They describe greater
exposure during longer generation, not an independent intervention on length.

The weight study shows the same length contrast with an unquantized cache:
math medians are 216, 278 and 249 tokens for Llama, Qwen and Mistral; QA
medians are 5, 4.5 and 7. Math changes remain more frequent within the first
64 tokens (80/256, 99/256, 114/256) than QA changes (7/128, 6/128, 14/128).
Rolling cache recompression therefore cannot by itself explain the pattern
across both studies. Three Mistral QA answers reach the cap; all other native
QA answers in the weight study reach EOS. Prompts, task content, answer format,
length and model behavior remain confounded. More generation offers more
positions at which changed numerical states can affect the next token, but
these records do not identify how much of the task difference is caused by
length or establish that document grounding itself protects consistency.
\end{document}